\documentclass[nohyperref]{article}

\usepackage{microtype}
\usepackage{graphicx}
\usepackage{subfigure}
\usepackage{booktabs} 

\usepackage{hyperref}

\usepackage[accepted]{icml2022}

\usepackage{amsmath}
\usepackage{amssymb}
\usepackage{mathtools}
\usepackage{amsthm}

\usepackage[capitalize,noabbrev]{cleveref}

\usepackage{array}
    \newcolumntype{P}[1]{>{\centering\arraybackslash}p{#1}}
    \newcolumntype{M}[1]{>{\centering\arraybackslash}m{#1}}
\usepackage{bbm}

\theoremstyle{plain}

\theoremstyle{definition}

\theoremstyle{remark}

\usepackage[textsize=tiny]{todonotes}

\icmltitlerunning{Policy Distillation with Selective Input Gradient Regularization for Efficient Interpretability}

\begin{document}

\twocolumn[
\icmltitle{Policy Distillation with Selective Input Gradient Regularization for Efficient Interpretability}



\icmlsetsymbol{equal}{*}

\begin{icmlauthorlist}
\icmlauthor{Jinwei Xing}{cog}
\icmlauthor{Takashi Nagata}{cs}
\icmlauthor{Xinyun Zou}{cs}
\icmlauthor{Emre Neftci}{cog}
\icmlauthor{Jeffrey L. Krichmar}{cog,cs}
\end{icmlauthorlist}

\icmlaffiliation{cog}{Department of Cognitive Sciences, University of California, Irvine, US}
\icmlaffiliation{cs}{Department of Computer Science, University of California, Irvine, US}

\icmlcorrespondingauthor{Jinwei Xing}{jinweix1@uci.edu}

\icmlkeywords{Machine Learning, ICML}

\vskip 0.3in
]



\printAffiliationsAndNotice{} 

\begin{abstract}
Although deep Reinforcement Learning (RL) has proven successful in a wide range of tasks, one challenge it faces is interpretability when applied to real-world problems. Saliency maps are frequently used to provide interpretability for deep neural networks. However, in the RL domain, existing saliency map approaches are either computationally expensive and thus cannot satisfy the real-time requirement of real-world scenarios or cannot produce interpretable saliency maps for RL policies. In this work, we propose an approach of Distillation with selective Input Gradient Regularization (DIGR) which uses policy distillation and input gradient regularization to produce new policies that achieve both high interpretability and computation efficiency in generating saliency maps. Our approach is also found to improve the robustness of RL policies to multiple adversarial attacks. We conduct experiments on three tasks, MiniGrid (Fetch Object), Atari (Breakout) and CARLA Autonomous Driving, to demonstrate the importance and effectiveness of our approach. 
\end{abstract}

\section{Introduction}
\label{submission}
Reinforcement learning (RL) systems have achieved impressive performance in a wide range of simulated domains such as games \citep{mnih2015human, silver2016mastering, vinyals2019grandmaster} and robotics \citep{lillicrap2015continuous, fujimoto2018addressing, haarnoja2018soft}. However, the interpretability of an agent's decision making and robustness to attacks need to be addressed when applying RL to real-world problems. For instance, in a self-driving scenario, real-time interpretability could explain how an RL agent produces a decision in response to its observed states and enable a safer deployment under real-world conditions and adversarial attacks \citep{ferdowsi2018robust}.

Saliency maps in deep learning is a technique used to interpret input features that are believed to be important for the neural network output \citep{simonyan2013deep, selvaraju2017grad, fong2017interpretable, smilkov2017smoothgrad, sundararajan2017axiomatic, zhang2018top}. As the issue of interpretability in RL gets more attention, a number of methods have been proposed to generate saliency maps to explain the decision making of RL agents. Existing saliency map methods in RL either use gradients to estimate the influence of input features on the output \citep{wang2016dueling} (gradient-based methods) or compute the saliency of an input feature by perturbing it and observing the change in output \citep{greydanus2018visualizing, iyer2018transparency, Puri2020Explain} (perturbation-based methods). Gradient-based methods can compute saliency maps efficiently with backpropagation. However, the quality of gradient-based saliency maps is generally poor \cite{rosynski2020are}.  Perturbation-based methods are effective in highlighting the important features of the input, but at a significant computationally cost, which can make them ineffective when deployed on systems with real-time constraints. As a result, existing RL agents cannot provide high interpretability in a computation-efficient manner.

Different from previous work proposing new saliency calculation methods, we focus on improving the natural interpretability of RL policies. Given a RL policy, we propose an approach of Distillation with selective Input Gradient Regularization (DIGR) that uses policy distillation and input gradient regularization to retrain a new policy. In our approach, input gradient regularization selectively regularizes gradient-based saliency maps of the policy to imitate its interpretable perturbation-based saliency maps. This allows the new RL policy to generate high-quality saliency maps with gradient-based methods and thus achieve both high interpretability and computational efficiency. At the same time, to ensure that input gradient regularization does not cause task performance degradation, we use policy distillation \cite{czarnecki2019distilling} to constrain the output of the new RL policy to remain close to the original RL policy.


We evaluate our method in three different tasks, which include an object fetching task from MiniGrid \cite{gym_minigrid}, Breakout from Atari games and CARLA Autonomoud Driving \cite{dosovitskiy2017carla}. The results show that RL policies trained with our approach are able to achieve efficient interpretability while maintaining good task performance. Selective input gradient regularization also improves the robustness of RL policies to adversarial attacks. These two desired properties allow the RL policy to better adapt to real-world scenarios.

To summarize, we demonstrate a novel approach to improve the efficient interpretability and robustness to attacks of RL policies based on the utilization of saliency maps. Our approach increases the applicability of RL to real-world problems.

\section{Background and Motivation}

\begin{figure*}[h!]%
\centering
\subfigure[Saliency Maps of Red-Fetch-Green]{%
\label{fig:motivation_interpretability}%
\includegraphics[width=0.63\textwidth]{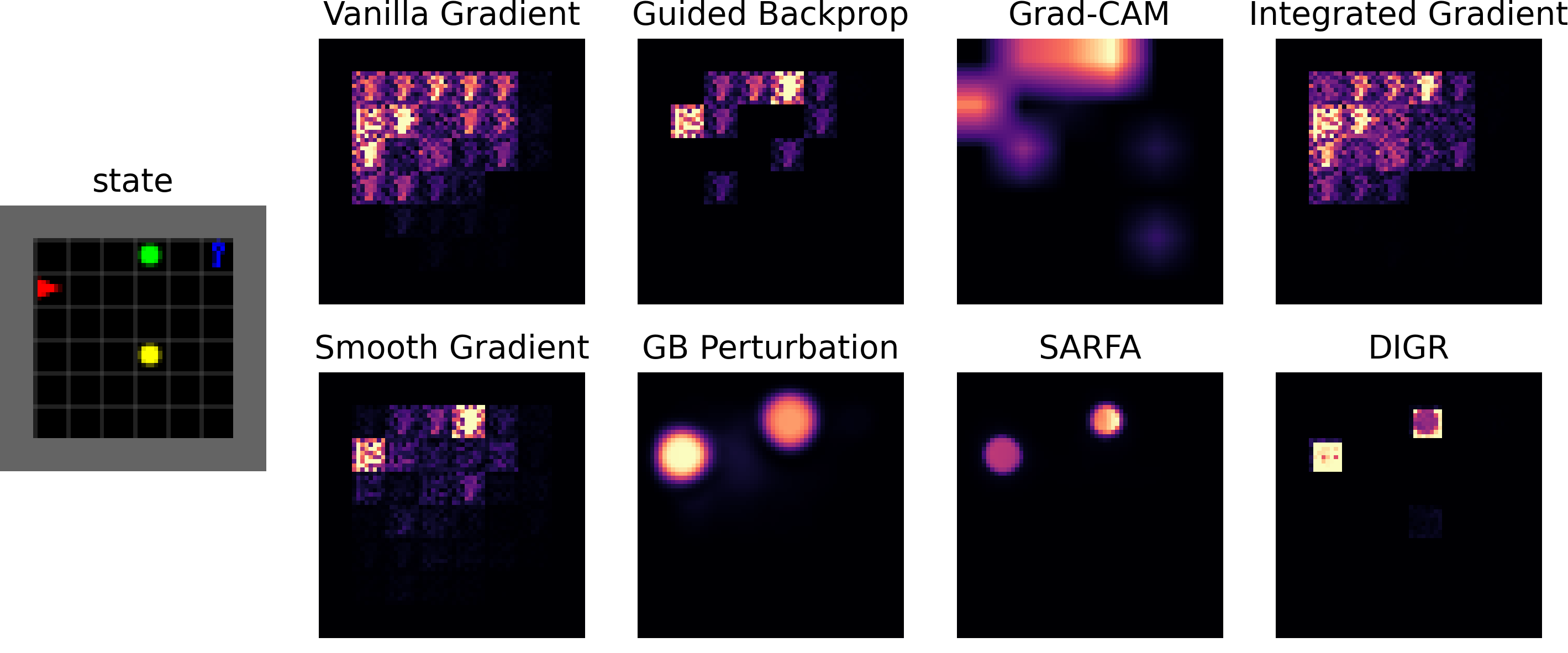}}%
\qquad
\subfigure[Saliency Map Generation Time]{%
\label{fig:motivation_efficiency}%
\includegraphics[width=0.32\textwidth]{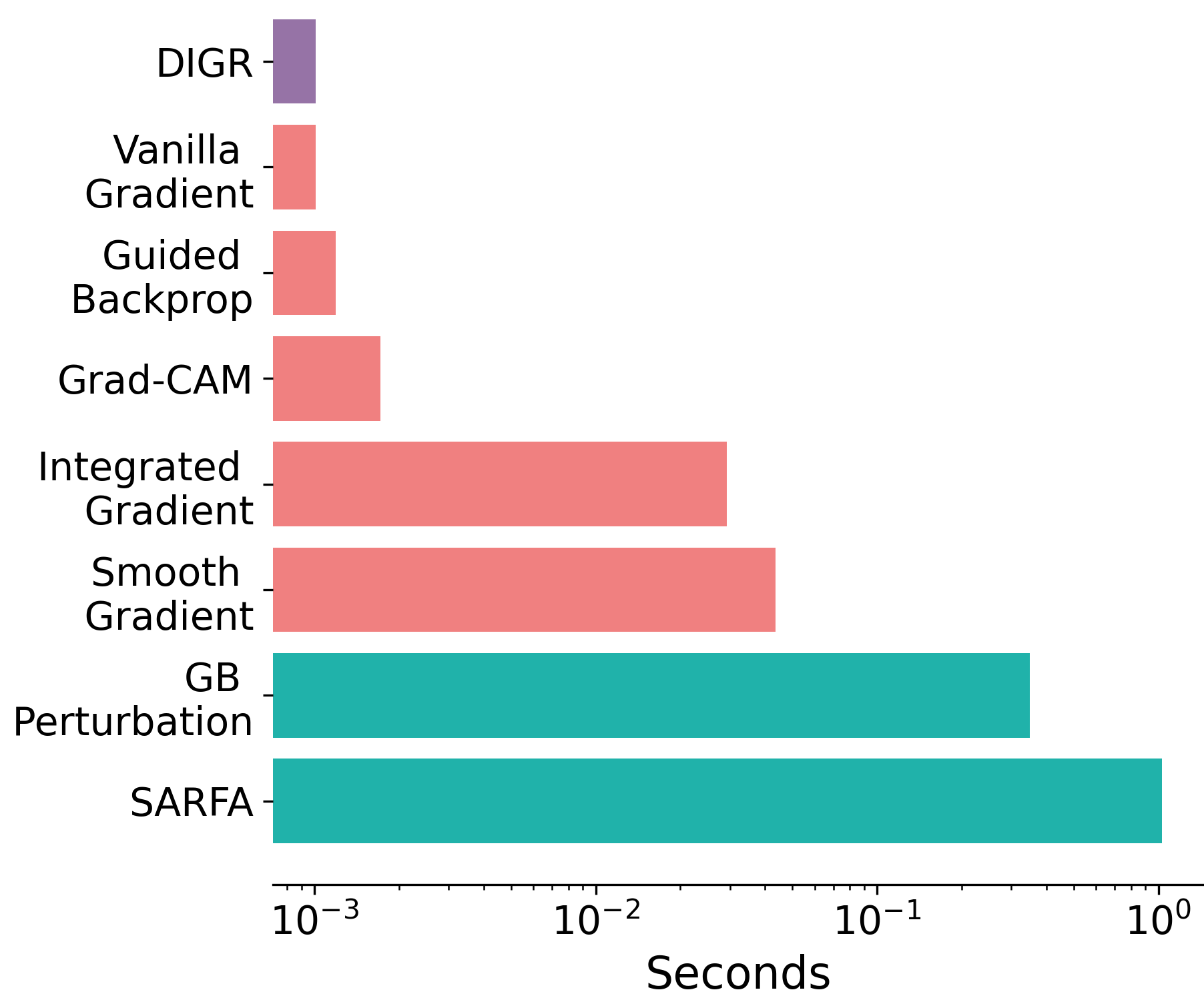}}%
\caption{(a). Different saliency maps on Red-Fecth-Green. All gradient-based saliency maps (Vanilla Gradient, Guided Backprop, Grad-CAM, Integrated Gradient and Smooth Gradient) produced by the PPO policy are noisy and show noticeable saliency on task-unrelated features. Gaussian-Blur Perturbation (GB Perturbation), SARFA saliency maps and saliency maps produced by DIGR approach demonstrate saliency on the red agent and green target object only. (b). The average time for each method to generate one saliency map for states of Red-Fetch-Green during policy deployment with a CPU of Intel i7-9750H and a GPU of GeForce RTX 2080 Ti. We mark DIGR with purple and use red and green colors to represent normal gradient-based and perturbation-based saliency map methods.}
\label{fig:motivation}
\end{figure*}

\paragraph{Reinforcement Learning} In reinforcement learning, agents learn to take actions in an environment that maximize their cumulative rewards. The environment is typically stated in the form of a Markov Decision Process (MDP), which is expressed in terms of the tuple ($S, A, T, R$) where $S$ is the state space, $A$ is the action space, $T$ is the transition function and $R$ is the reward function. At each time step $t$ in the MDP, the agent takes an action $a_t$ in the environment based on current state $s_t$ and receives a reward $r_{t+1}$ and next state $s_{t+1}$. The goal of the agent is to find a policy $\pi(s)$ to select actions that maximize the discounted cumulative future rewards $r_t + \gamma r_{t+1} + \gamma^2 r_{t+2} + ...$, where $\gamma$ is the discount factor ranging from $0$ to $1$.   

\paragraph{Policy Distillation}
Policy distillation \cite{rusu2015policy, czarnecki2019distilling} transfers knowledge from one teacher policy $\pi_{t}$ to a student policy $\pi_{s}$ by training the student policy to produce the same behavior as the teacher policy. This is normally achieved by supervised regression to minimize the following objective: 
\begin{equation}
    \mathop{J} = \mathbb{E}_{s \sim \pi_c} [D(\pi_t(s), \pi_s(s))],
\end{equation}
where $\pi_c$ is the control policy that interacts with the environment to produce states for training, and $D$ is a distance metric. There are multiple choices for both $\pi_c$ and $D$. For example, the control policy $\pi_c$ could take the form of the teacher policy $\pi_t$ or student policy $\pi_s$ or even a combination of them. Suitable distance metrics could be mean squared error or KL divergence.

\paragraph{Saliency Map in RL}
Saliency map techniques are popular in computer vision and RL communities for interpreting deep neural networks. Gradient-based methods calculate the gradient of some function $f$ with respect to inputs $s$ based on the chain rule and then use the gradients to estimate the influence of input features on the output. In RL, one common approach is the Jacobian saliency map \citep{wang2016dueling} which computes the saliency of input feature $s_i$ as $\lvert \frac{\partial f(s)}{\partial s_i} \rvert$ where function $f$ could be calculated from either the state-action value $Q(s, a)$ in Q-learning or the action distribution $\pi(s)$ in actor-critic methods. Other gradient-based visualization methods from the field of image classification are also explored \cite{greydanus2018visualizing, rosynski2020are} but most of them didn't work well in the RL domain.  

Perturbation-based methods compute the saliency of an input feature by perturbing (e.g. removing, altering or masking) the feature and observing the change in output. Given a state input $s$, a perturbed state $s'$ could be generated by inducing a perturbation on input feature $s_i$. The approach of computing the change in output caused by the perturbation may vary based the form of RL agent.  For example,  in Q-learning, the network output is a scalar and thus the saliency of $s_i$ could be defined as $|Q(s,a) - Q(s',a)|$. In actor-critic methods, the saliency of $s_i$ could be defined as $D_{KL} (\pi(s) || \pi(s'))$ which is the KL divergence between action distributions before and after the perturbation. Alternatively, \cite{greydanus2018visualizing} considers the output of actor as a vector and computes the saliency as $\frac{1}{2} ||\pi(s) - \pi(s')||^2 $. \citet{Puri2020Explain} further proposed an approach called SARFA to addresses the specificity and relevance in perturbation-based saliency maps.

\paragraph{Motivation}
We first introduce a simple fetching-object task in MiniGrid and demonstrate the results of different saliency map methods on this task to motivate our method. In the fetching-object task in MiniGrid, the environment is a room composed of 8x8 grids and 4 entities with unique colors. The red agent needs to locate and pick up the green object, while the yellow and blue objects are distractors. Based on the task rule, we name this task as Red-Fetch-Green. We first use PPO \cite{schulman2017proximal} to train a RL policy to solve the task and then investigate the interpretability and computation efficiency of different saliency map methods to explain the policy. Examples of gradient-based (Vanilla Gradient \cite{simonyan2013deep}, Guided Backprop \cite{springenberg2014striving}, Grad-CAM  \cite{selvaraju2017grad}, Integrated Gradient \cite{sundararajan2017axiomatic}, Smooth Gradient \cite{smilkov2017smoothgrad}) and perturbation-based (Gaussian-Blur Perturbation \cite{greydanus2018visualizing} and SARFA \cite{Puri2020Explain}) saliency maps for Red-Fetch-Green are shown in Figure \ref{fig:motivation_interpretability}. We also include an example of saliency map generated by our DIGR approach for comparison. In general, for methods except DIGR, perturbation-based saliency maps mainly demonstrate high saliency on task relevant features (e.g. red agent and green target object) while gradient-based saliency maps are more noisy and harder to interpret. However, the high quality of perturbation-based saliency maps are achieved with an increased cost of computation time as shown in Figure \ref{fig:motivation_efficiency}. The computation time of perturbation-based saliency maps are highly affected by the input size and policy network architectures. This makes it incompatible with many real-world tasks that require real-time interpretability such as autonomous driving. Thus, based on the result in Figure \ref{fig:motivation}, we find that normal gradient-based saliency maps are computationally more efficient but hard to interpret while perturbation-based saliency maps are more interpretable but come with a higher computation cost during deployment. This finding motivates us to think about how we can keep the computation efficiency of gradient-based methods and high interpretability of perturbation-based methods while avoiding their limitations, and thus propose DIGR. 

How does DIGR generate interpretable saliency maps like perturbation-based methods while only requiring a short generation time as the most efficient Vanilla Gradient saliency maps? Is it possible for us to use gradient-based methods such as Vanilla Gradient method to generate high-quality saliency maps as those from perturbation-based methods? We answer these questions in the next section.

\section{Method}
Our approach to achieve both computational efficiency and high interpretability in RL is to produce a policy whose gradient-based saliency maps are comparable to those of perturbation-based methods. To achieve this, given a trained RL policy, we set its perturbation-based saliency maps as supervisory signals and update the weights of the policy so that its gradient-based saliency maps match the perturbation-based saliency maps. Since the computations involved in gradient-based saliency maps are differentiable, we can use stochastic gradient descent to conduct the training. The idea of optimizing gradient-based saliency maps has a close connection with input gradient regularization which imposes constraints on how input gradients behaves. 
For example, \citet{ross2018improving} penalizes input gradients based on expert annotation to prevent the network from ``attending'' to certain parts of the input in an image classification task. Inspired by this, the training of the gradient-based saliency map in our approach is conducted by selectively penalizing the gradients of input features that have low perturbation-based saliency.

One challenge of selective input gradient regularization is that optimizing gradient-based saliency maps may also affect the policy output and thus degrade the task performance. To avoid this, we conduct policy distillation to ensure that the new policy maintains the same task performance. We give a more formal introduction of our method below.

Given a RL policy $\pi$ and input $s$, we define the function $g$ as the method used in generating gradient-based saliency map $M_g$ and function $f$ as the method used in generating perturbation-based saliency map $M_p$. Both $M_g$ and $M_p$ have the same size as input $s$. Each element in the saliency map, ${M_g}_i$ and ${M_p}_i$, are computed as
\begin{equation}
\begin{aligned}
g(s, i, \pi) &=  \mid \sum_a \pi(a|s) \frac{\partial \pi(a|s)}{\partial s_i} \mid   \\   {M_g}_i  &= \frac{g(s,i,\pi)}{\max\limits_{0 \leq j \leq N} g(s, j,\pi) } 
\end{aligned}
\end{equation}

\begin{equation}
\begin{aligned}
f(s, i, \pi) &= D_{KL}(\pi(s)|| \pi(m(s, i)))  \\
{M_p}_i  &= \frac{ f(s,i,\pi) }{\max\limits_{0 \leq j \leq N}  f(s, j,\pi)}
\end{aligned}
\end{equation}

\begin{figure*}[h]
    \centering
    \includegraphics[width=0.8\textwidth]{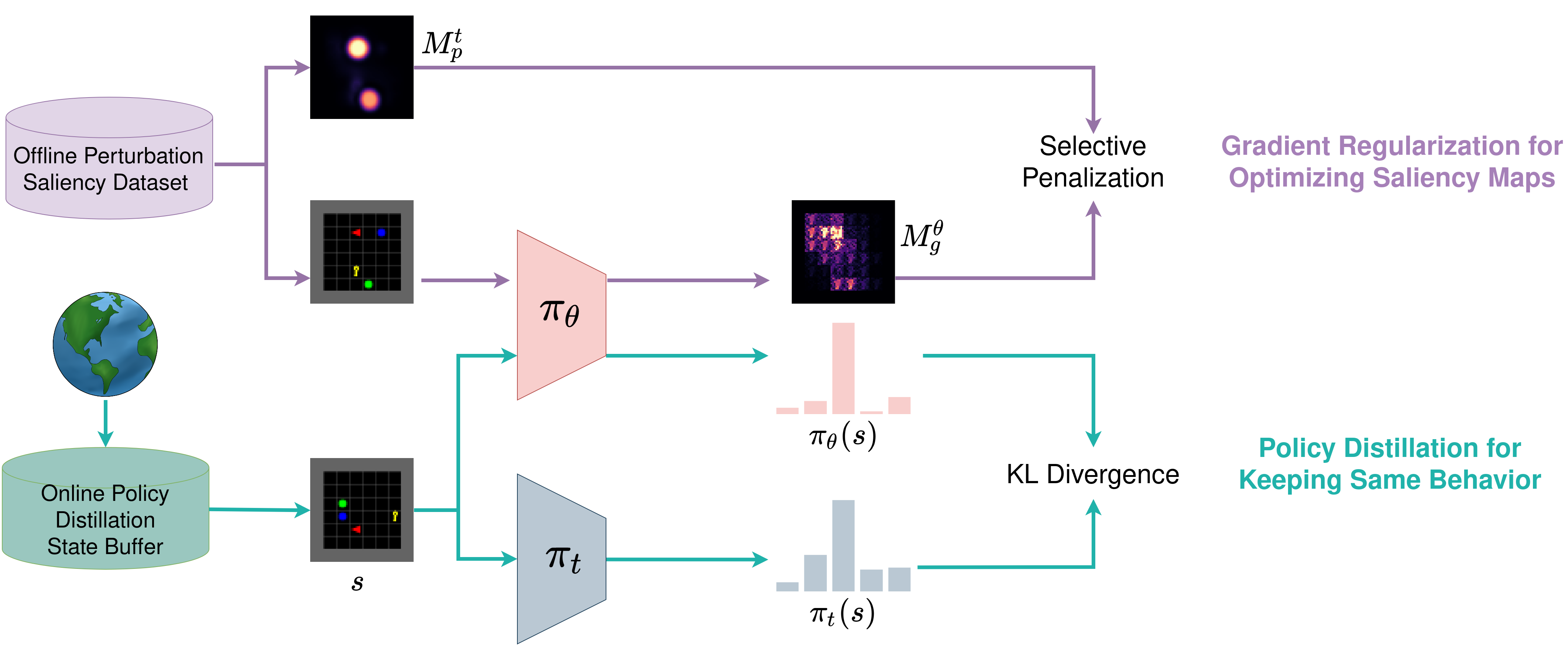}
    \vspace*{-3mm}
    \caption{Framework of our approach. Policy $\pi_\theta$ is used as the control policy and interact with the environment. The experienced states are saved into a replay buffer and then sampled later for policy distillation. The training includes two objectives. The first objective is using input gradient regularization to regularize gradient-based saliency map $M_g^\theta$ based on the perturbation-based saliency map $M_p^t$. The second objective is using policy distillation to make sure the learning policy $\pi_\theta$ have the same behavior as the trained policy $\pi_t$.}
    \label{fig:framework}
\end{figure*}

where $g(s, i, \pi)$ and $f(s, i, \pi)$ compute the gradient-based and perturbation-based saliency values of input feature $s_i$ given policy $\pi$. These saliency values are then normalized between 0 and 1 to form saliency maps which contain $N$ elements in each map. In this work, perturbation function $m$ induces a Gaussian blur on the input with the input feature of interest $s_i$ as the center \cite{greydanus2018visualizing}. It's worth mentioning that, besides perturbation-based saliency maps, DIGR could be easily extended to utilize other saliency data (e.g. saliency maps from expert annotation) as supervisory signals. In this work, we focus on using perturbation-based saliency maps for input gradient regularization as they show high interpretability and can be computed as long as we have access to the policy and states.


After introducing the process of generating two types of saliency maps given a RL policy and state input, we introduce how they are used in DIGR. Given a trained RL policy $\pi_t$, DIGR aims to produce a new policy $\pi_\theta$ with parameters $\theta$ that can generate interpretable saliency maps using gradient-based method. Given a state input $s$, the saliency map could differ based on the generation method (gradient-based vs perturbation-based) and the policy ($\pi_t$ vs $\pi_\theta$) used to generate them. For clarity, we define these 4 types of saliency maps as ${M_g}^{t}$,  ${M_g}^{\theta}$, ${M_p}^{t}$,  ${M_p}^{\theta}$ where $M_g$ and $M_p$ represent gradient-based and perturbation-based saliency maps, respectively. The superscripts $t$ and $\theta$ represent whether the saliency map is generated by the given trained policy $\pi_t$ or the new policy $\pi_\theta$. Then the loss function for input gradient regularization is
\begin{equation}
\begin{aligned}
    L =  \mathbb{E}_{s \sim d_{\pi_\theta}}[\frac{1}{N}  \sum_{i=1}^{N} {\mathbbm{1}_{[0, \infty)}(\lambda - {M_p}^t_i) \times {M_g}^{\theta}_i }]
\end{aligned}
\end{equation}

where $d_{\pi_\theta}$ is the state distribution following policy $\pi_\theta$ and $N$ is the number of input features in the saliency map. ${M_p}^t$ and ${M_g}^\theta$ have the same size and are both indexed by $i$. Threshold $\lambda$ is used in the indicator function $\mathbbm{1}$ to determine whether one input gradient should be penalized. The indicator function $\mathbbm{1}$  returns 1 if $\lambda - {M_p}^t_i \geq 0$ and 0 otherwise. In other words, if the perturbation-based saliency for an input feature is below threshold $\lambda$, the loss penalizes its gradient-based saliency. This selective penalization allows the model to only keep high saliency on task-relevant features selected by the perturbation-based saliency maps.

\begin{figure*}[h!]
    \centering
    \includegraphics[width=0.9\textwidth]{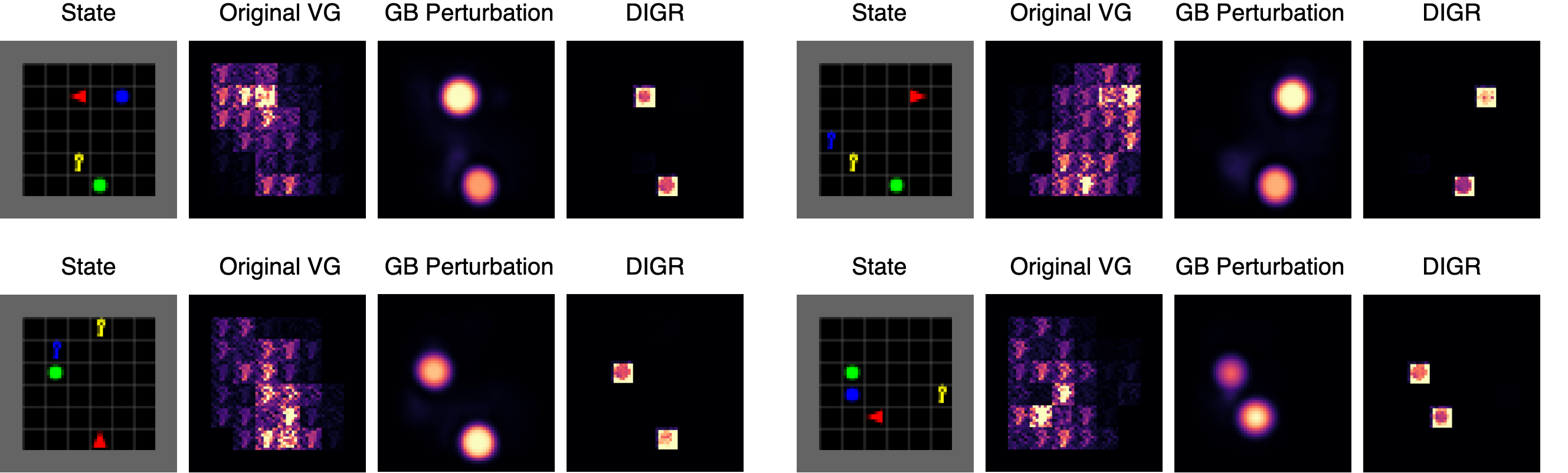}
    \caption{Demonstration of our approach on Red-Fetch-Green. There are four sets of examples and each set includes a state, a Vanilla Gradient saliency map generated by the original policy (Original VG), a Gaussian-Blur perturbation-based saliency map (GB Perturbation) generated by the original policy and a Vanilla Gradient saliency map generated by the policy trained with DIGR. The annotation of DIGR on the figure refers to Vanilla Gradient saliency maps generated by the policy trained with DIGR. In all examples, GB Perturbation and DIGR saliency maps show high saliency on the red agent and green target while Original VG saliency maps are noisy and hard to interpret.}
    \label{fig:demo_minigrid}
\end{figure*}

The final loss function in our approach is a weighted combination of selective input gradient regularization and policy distillation. In practice, generating perturbation-based saliency maps online for input gradient regularization could be time-consuming and slow down the overall training. To address this, we build an offline perturbation saliency dataset $D$ which contains states sampled from $d_{\pi_t}$ and the corresponding perturbation-based saliency maps generated in advance. Because of the policy similarity brought by policy distillation, we use $D$ to approximate $d_{\pi_\theta}$ for input gradient regularization. As a result, the loss function for DIGR is 

\vspace*{-4mm}
\begin{equation}
\begin{aligned}
L_{DIGR} = \mathbb{E}_{s \sim D} [\underbrace{\frac{1}{N}  \sum_{i=1}^{N} { \mathbbm{1}_{[0, \infty)}(\lambda - {M_p}^t_i) \times {M_g}^\theta_i}}_\textrm{Input Gradient Regularization}] \\ +  \alpha \mathbb{E}_{s \sim d_{\pi_\theta}} [\underbrace{ \vphantom{ \sum_{i=1}^{N} } D_{KL}(\pi_t(s) || \pi_\theta(s))}_\textrm{Policy Distillation}]
\end{aligned}
\end{equation}

where $\alpha$ is a weighting parameter used to balance the loss of input gradient regularization and policy distillation. We show the complete architecture of our approach in Figure \ref{fig:framework}.

\section{Experimental Results}
We conducted experiments on three tasks including Red-Fetch-Green in MiniGrid, Breakout in Atari games and CARLA Autonomous Driving to demonstrate the effectiveness of our approach. 
In Red-Fetch-Green , the red agent needs to locate and pick up the green object while avoiding picking up other distractors in a room composed of 8x8 grids. In Breakout, the paddle is controlled to move at the bottom to ricochet the ball against the bricks and eliminate them for rewards. Besides these two tasks, we designed a CARLA Autonomous Driving task in which the agent needs to control an autonomous car driving on a highway while avoiding collisions. Since CARLA simulator has its simulation clock and time that can be matched with real time, we use it to demonstrate that both high quality and computation efficiency of our approach in interpreting RL policies are important in real-world scenarios.

\subsection{Setup}

\paragraph{RL Training} In our experiments, we first use PPO algorithm to train RL policies on Red-Fetch-Green, Breakout and CARLA Autonomous Driving. The trained RL policies, which are used to generate offline perturbation saliency datasets for input gradient regularization, also serve as the teacher policy in policy distillation and generate saliency maps for comparison. In all three tasks, we used similar network architectures composed of 3 convolutional layers and 2 linear layers but with different layer sizes. The trained RL policies achieved reasonable good performance in each task: The policy in Red-Fetch-Green solves the task with a success rate of 100\%; the policy in Breakout achieves an average score of 320; the policy in CARLA Autonomous Driving could drive smoothly and learned to steer to avoid collision with other vehicles. We include more details of RL training in the appendix.

\paragraph{Offline Perturbation Saliency Dataset} 
To conduct selective input gradient regularization, we generate an offline perturbation saliency dataset by sampling states experienced by the trained RL policy $\pi_t$ and generating the corresponding Gaussian-Blur perturbation saliency maps \cite{greydanus2018visualizing}. The perturbation saliency datasets of Red-Fetch-Green, Breakout and CARLA Autonomous Driving contain 1k, 10k, 2.5k pairs of states and saliency maps. Although our method still needs to generate perturbation-based saliency maps, the computation happens in the training stage without affecting the computation efficiency during deployment. Also, the computation problem could be mitigated by the limited size of the dataset (e.g. 1k, 10k and 2.5k states in Red-Fetch-Green, Breakout, and CARLA respectively) and the potential utilization of parallel computing with multiple machines.

\paragraph{DIGR Training} 
DIGR uses selective input gradient regularization and policy distillation to produce a new policy that achieves efficient interpretability while maintaining task performance. In all three experiments, we randomly initiate the new policy $\pi_\theta$. To further stabilize the training, we consider the training of selective input gradient regularization and policy distillation as a multi-objective optimization problem and used the technique of PCGrad \cite{NEURIPS2020_3fe78a8a} to mitigate gradient interference. More hyperparameters of training are included in the appendix.

\subsection{Effectiveness via Visual Illustrative Examples}

\begin{figure*}[h!]
    \centering
    \includegraphics[width=0.9\textwidth]{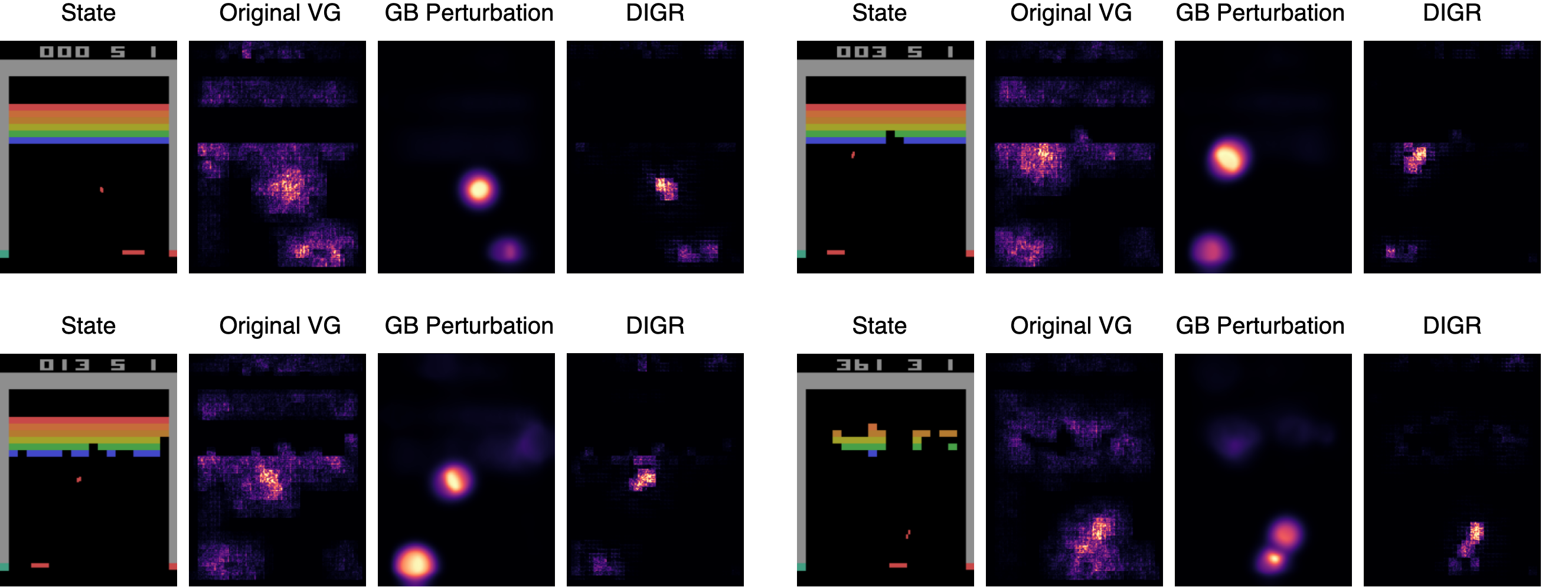}
    \caption{Demonstration of our approach on Breakout. VG and GB Perturbation stand for Vanilla Gradient and Gaussian-Blur Perturbation. Both DIGR and Gaussian-Blur perturbation-based saliency maps demonstrate high saliency mainly on the paddle and ball while the Vanilla Gradient saliency maps generated by the original policy (Original VG) are more noisy.}
    \label{fig:demo_breakout}
\end{figure*}

\begin{figure*}[h!]
    \centering
    \includegraphics[width=0.9\textwidth]{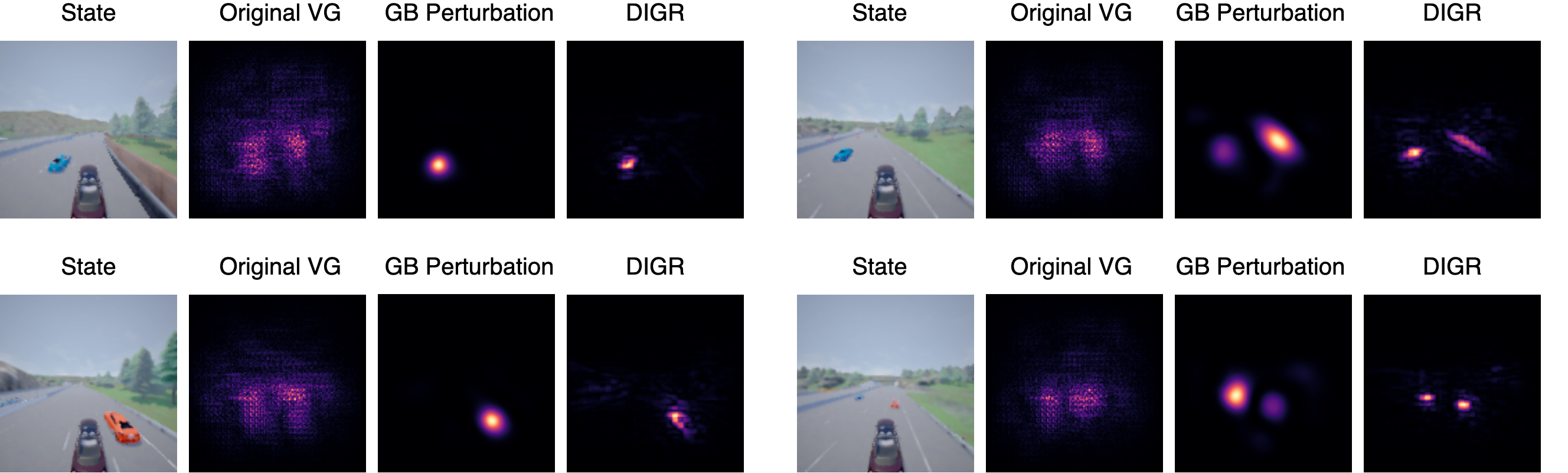}
    \caption{Demonstration of our approach on CARLA Autonomous Driving. VG and GB Perturbation stand for Vanilla Gradient and Gaussian-Blur Perturbation. In the left two sets of examples, DIGR and GB Perturbation methods demonstrate high saliency on the vehicles that got close to the controlled vehicle. In the top-right example, DIGR and GB perturbation methods show high saliency on the vehicle and road curb. In the bottom-right example, DIGR and GB perturbation methods show high saliency on two vehicles ahead. DIGR and GB perturbation methods didn't show saliency on the controlled vehicle because the controlled vehicle is always at the same region of the images for all states and is not salient to the performance. The saliency are demonstrated on other features that may lead to collision and affect the performance. In all four sets of examples, Vanilla Gradient saliency maps generated by the original policy (Original VG) are very similar and hard to distinguish.}
    \label{fig:demo_carla}
\end{figure*}

The main goal of our approach is allowing RL policies to generate interpretable saliency maps with computationally efficient gradient-based methods. To demonstrate the effectiveness of our approach, we provide examples of the most computationally-efficient Vanilla Gradient saliency maps before and after our method, and Gaussian-Blur perturbation saliency maps that work as supervisory guidance in Figure \ref{fig:demo_minigrid}, \ref{fig:demo_breakout} and \ref{fig:demo_carla}. 

Our results show that Vanilla Gradient saliency maps generated by original RL policies are noisy and hard to interpret. However, after the optimization with our approach, we can use the same saliency map method to generate much more interpretable saliency maps which reduces a large amount of unexplainable saliency and demonstrate high saliency on task-relevant features only. The saliency maps generated by our approach also have a close similarity to Gaussian-Blur perturbation-based saliency maps which demonstrates the successful saliency guidance. We provide more visual examples containing saliency maps produced by other gradient-based methods for comparison in the appendix.

\subsection{Importance of Computational Efficiency}
\begin{figure}
    \centering
    \includegraphics[width=1\linewidth]{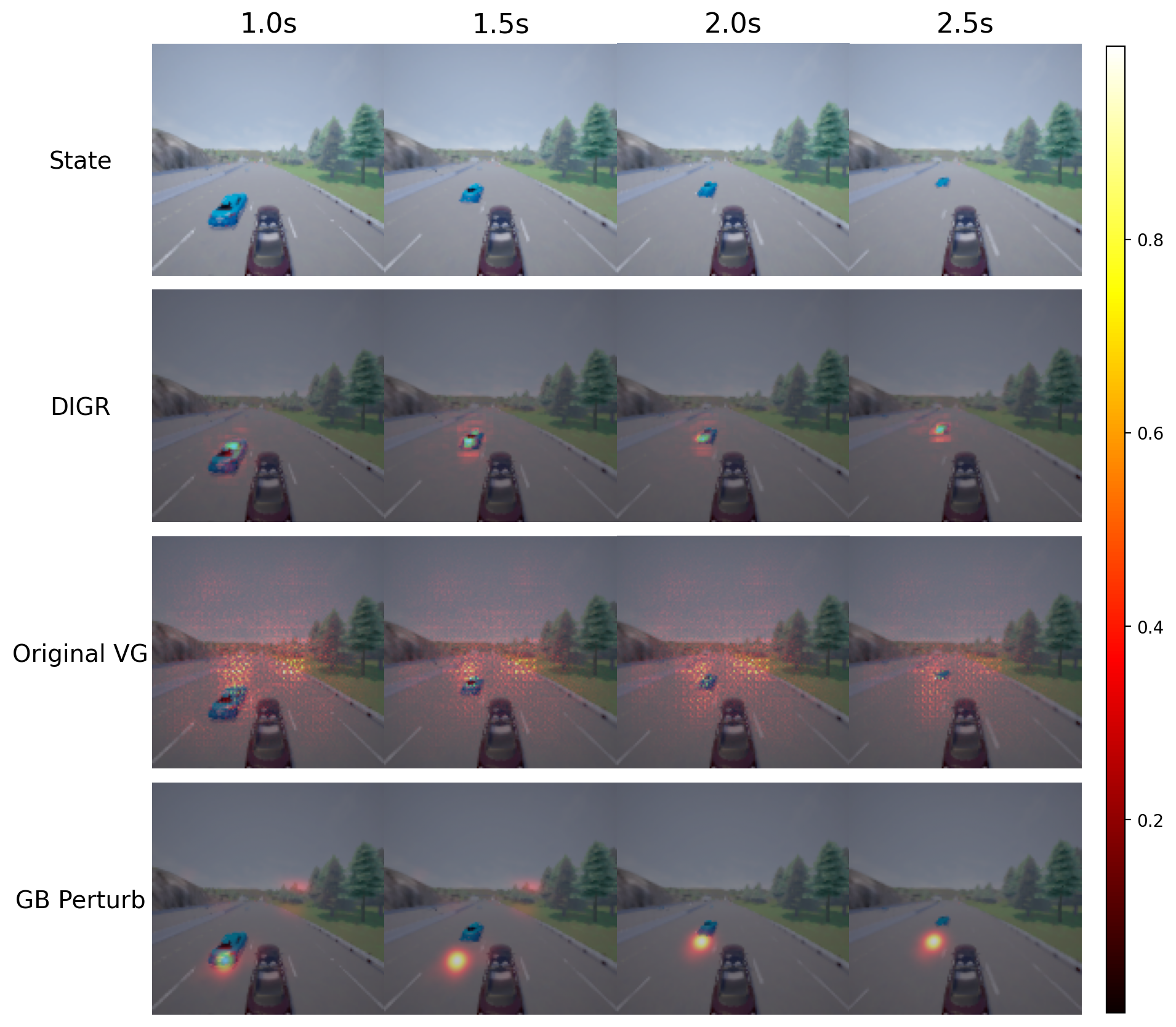}
    \vspace*{-3mm}
    \caption{Different types of saliency maps on a sequence of states in CARLA Driving. Vanilla Gradient saliency maps generated by the policy trained with DIGR always demonstrate high saliency on the traffic vehicles while Vanilla Gradient saliency maps generated by the original policy (original VG) are noisy and just show saliency in the center region of all states. Gassuain-Blur perturbation-based saliency maps show saliency behind the vehicle because of the computation delay. The bar on the right represents the mapping between saliency values and colors.}
    \label{fig:carla_sequence_demo}
\end{figure}

In this section, we further show the importance our approach by demonstrating that missing either computation efficiency or high interpretability make it difficult to achieve interpretable RL in real-world scenarios. We take Autonomous Driving as an example and show the results of utilizing different saliency maps to explain a sequence of RL decision making in Figure \ref{fig:carla_sequence_demo}. In our experiments, the state of CARLA Autonomous Driving is a 128x128 RGB image taken every 0.05 seconds by a camera attached to the ego vehicle. Although Gaussian-Blur perturbation-based saliency maps show high interpretability as seen in Figure \ref{fig:demo_carla}, it takes 0.97±0.02 seconds to generate one saliency map with a GPU of RTX 2080Ti. This means there's a delay of almost one second between meeting the state and the availability of corresponding saliency map and all saliency maps for states experienced during the delay will be missed. In constrast to Gaussian-Blur perturbation-based saliency maps that each takes 0.97 seconds to generate in average, Vanilla Gradient saliency maps are much more efficient to compute and take only 0.0021±0.0001 seconds for each state with the same machine. However, Vanilla Gradient saliency maps generated by normal RL policies are hard to interpret and only our approach achieves both computation efficiency and high interpretability.

\subsection{Saliency Dataset and Evaluation}
\begin{figure}[t]%
\centering
\subfigure[state]{%
\label{fig:dataset_state}%
\includegraphics[width=0.3\linewidth]{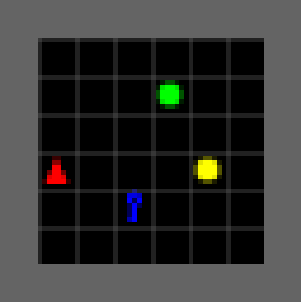}}%
\quad
\subfigure[important saliency]{%
\label{fig:dataset_saliency}%
\includegraphics[width=0.3\linewidth]{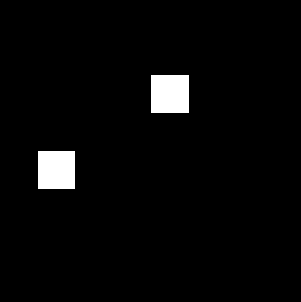}}%
\quad
\subfigure[unimportant saliency]{%
\label{fig:dataset_saliency}%
\includegraphics[width=0.3\linewidth]{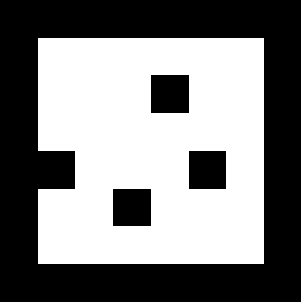}}%
\caption{\textbf{a}. An example state in the saliency dataset of Red-Fetch-Green. \textbf{b}. Regions whose saliency are important. \textbf{c}. Regions whose saliency are unimportant.}
\label{fig:dataset}
\end{figure}

Besides illustrative examples, we also aim to provide a quantitative evaluation of saliency maps generated by different approaches and thus introduce a new saliency dataset based on Red-Fetch-Green. Different from previous work that relies on expert annotations and classifies each state element as either important or unimportant feature \cite{Puri2020Explain}, we focus on features whose saliency importance are certain. There are six types of objects in Red-Fetch-Green including the red agent, the green target object, the blue and yellow distrators, grey walls and black empty grids. Based on the roles of objects, we assume the red agent and green target are important features as they have the most important information required for optimal decision making and assume the empty tiles as unimportant features since they do not provide any information. The two distractors and grey walls are not included in the dataset because their influence on decision making is either uncertain or only exists in a small subset of state space. We collected 10k states in the saliency dataset and provide an example in Figure \ref{fig:dataset}.

\begin{table}[h]
\centering
\begin{tabular}{@{}cccc@{}}
\toprule
      & \multicolumn{3}{c}{Saliency on Red-Fetch-Green} \\ 
                      \cmidrule(l){2-4} 
                      & important   & unimportant  & AUC     \\ \midrule
VG     & 56.04        & 278.10        & 0.840   \\
Guided BP & \textbf{82.84} & 35.67 & 0.993 \\
Grad-CAM & 43.12 & 364.97  & 0.686 \\
Smooth G & \textbf{83.05} & 84.76  & 0.991 \\
Integrated G & 67.79 & 232.09  & 0.900 \\
GB Perturbation & \textbf{86.11}         & 77.81    & 0.989   \\
SARFA &58.40 & 42.17 & 0.895 \\
\textbf{DIGR}                  & \textbf{72.52}         &  \textbf{0.00}        & \textbf{0.997}  \\ \bottomrule
\end{tabular}
\caption{Saliency results of Vanilla Gradient (VG), Guided Backpropagation (Guided BP), Grad-CAM, Smooth Gradient (Smooth G), Integrated Gradient (Integrated G), Gaussian-Blur Perturbation (GB Perturbation), SARFA of the original policy and Vanilla Gradient of DIGR policy on Red-Fetch-Green. Our method keeps comparable amount of important saliency, reduces all unimportant saliency and achieves the highest AUC.}
\label{tab:dataset_res}
\end{table}

\begin{figure*}[t]
    \centering
    \includegraphics[width=1\textwidth]{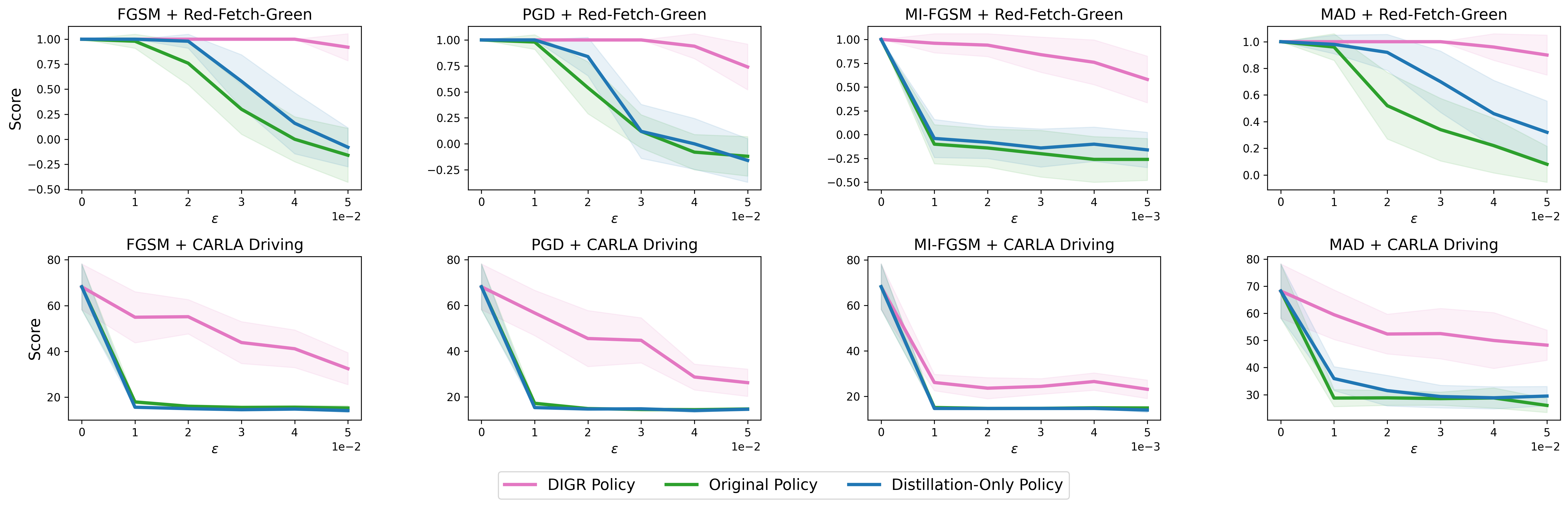}
    \vspace*{-3mm}
    \caption{Plicies trained with DIGR achieve much stronger robustness to all four types of adversarial attacks (FGSM, PGD, MI-FGSM and MAD) compared to the policies trained with normal RL algorithms. Although policy distillation also helps robustness slightly, selective input gradient regularization makes the most contribution to the improved robustness. All results are averaged over 50 runs in Red-Fetch-Green and 20 runs in CARLA Autonomous Driving. Shaded area represents one standard deviation. }
    \label{fig:robustness}
\end{figure*}

To evaluate the quality of different saliency maps, we compute the average amount of important saliency and unimportant saliency in each saliency map. Furthermore, we also compare different saliency maps with AUC, which is a popular metric used to evaluate saliency maps \citep{iyer2018transparency, Puri2020Explain}. As shown in Table \ref{tab:dataset_res}, our approach keeps a comparable amount of important saliency, reduces all unimportant saliency and achieves the highest AUC compared with other approaches. The decreased amount of unimportant saliency is in line with our expectation since our approach works by penalizing the saliency that are not helpful for interpretation. As a result, our approach utilizes gradient-based and perturbation-based saliency maps for training and finally achieves even better saliency maps.

\subsection{Policy Performance Maintenance}
\begin{figure}[h]
    \centering
    \includegraphics[width=0.95\linewidth]{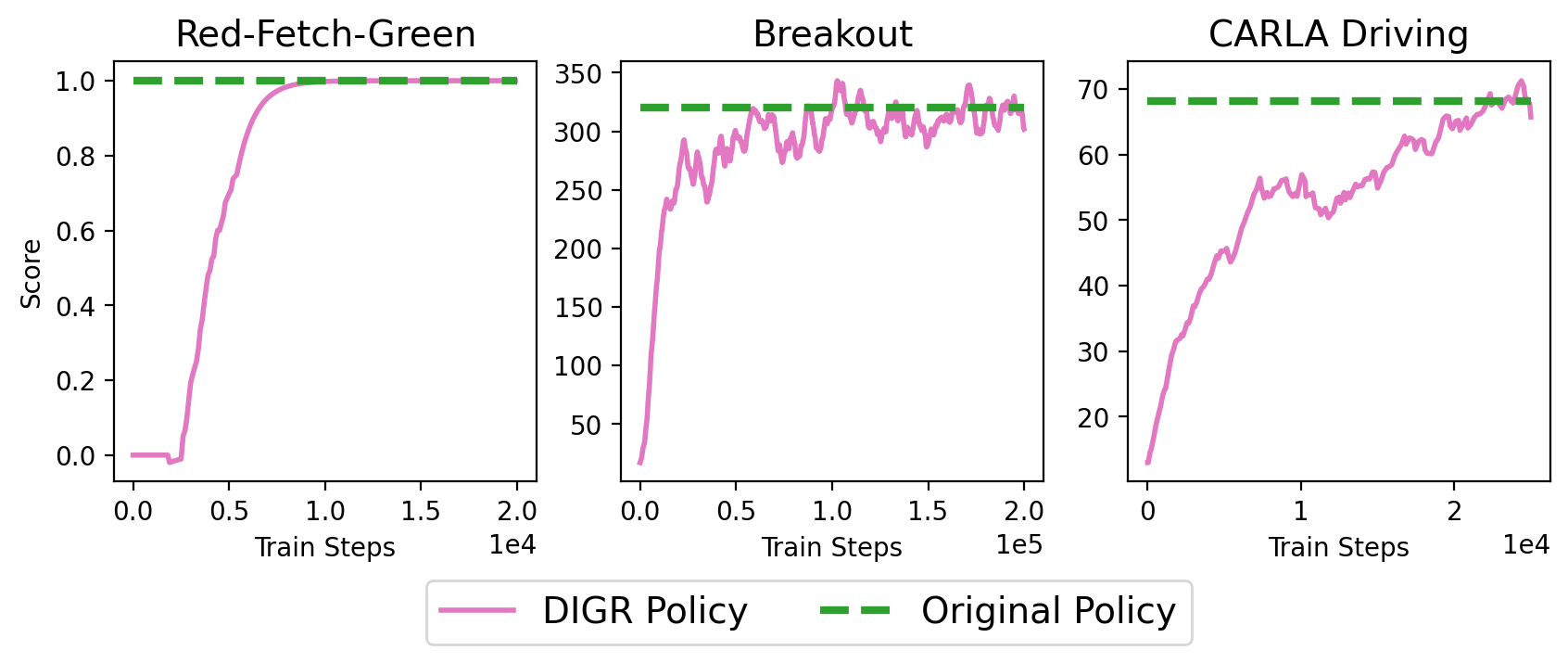}
    \caption{The performance of DIGR policy could match the performance of the original policy.}
    \label{fig:policyperformance}
\end{figure}

The objective of optimizing gradient-based saliency maps may change the action selection of the original policy and thus cause the policy performance to degrade. In DIGR, we use policy distillation to constrain the output of the new RL policy to remain close to the original policy. To verify its effectiveness, we plot the performance of DIGR policy during training and compare it with the results of the original policy. As seen in Figure \ref{fig:policyperformance}, the policy trained with our approach could achieve similar performance as the original policy.

\subsection{Improved Robustness to Attacks}

Recent research shows a deep entanglement between adversarial attacks and interpretability of deep neural network (DNN) models \citep{NEURIPS2018_b9946974, NEURIPS2019_7392ea4c}. Since DIGR improves the interpretability of Deep RL policies, we are also interested in its influence on policy's robustness to attacks. To study that, we evaluate the robustness of RL policies before and after applying DIGR to four types of adversarial attacks including Fast Gradient Sign Method (FGSM) \cite{huang2017adversarial}, Projected Gradient Descent (PGD) \cite{madry2018towards}, Momentum Iterative Fast Gradient Sign Method (MI-FGSM) \cite{dong2018boosting} and Maximum Action Difference (MAD) \citep{NEURIPS2020_f0eb6568} in Red-Fetch-Green and CARLA Autonomous Driving tasks. Since both policy distillation and input gradient regularization in our approach could affect the robustness of RL policies, we further include an ablation study by conducting policy distillation only to understand their own influence on robustness. As shown in Figure \ref{fig:robustness}, our approach significantly improves the robustness of RL policies. Although policy distillation also improves the robustness slightly, selective input gradient regularization contributes the most to the significant robustness gains.

\section{Conclusion}
We propose an approach called DIGR to improve the efficient interpretability of RL by retraining a policy with selective input gradient regularization and policy distillation. Our approach allows RL policies to generate highly interpretable saliency maps with computationally efficient gradient-based methods. We further show that our approach is able to improve the robustness of RL polices to multiple adversarial attacks. Interpretable decision-making and robustness to attacks are two challenges in deploying RL to real-world systems. We believe our approach could help to build trustworthy agents and benefit the deployment of RL policies in practice.

\nocite{langley00}

\bibliography{example_paper}
\bibliographystyle{icml2022}

\newpage
\appendix
\onecolumn

\section{Experiment details and hyperparameters}
We conduct experiments on three tasks including Red-Fetch-Green in MiniGrid, Breakout in Atari games and CARLA Autonomous Driving to demonstrate the effectiveness of our approach. To generate the trained reinforcement learning (RL) policies, we use Proximal Policy Optimization (PPO) as the trianing algorithm and list hyperparameters in Table \ref{tab:ppo}.

\begin{table*}[h]
\centering
\begin{tabular}{@{}cccc@{}}
\toprule
Hyperparameters            & Red-Fetch-Green & Breakout & CARLA Driving \\ \midrule
$\gamma$                      & 0.99        & 0.99     & 0.999     \\
$\lambda$                     & 0.95        & 0.95     & 0.95      \\
entropy bonus coefficient  & 0.01        & 0.01     & 0.01      \\
value loss coefficient     & 0.5         & 0.5      & 0.5       \\
gradient clipping          & 0.5         & 0.5      & 0.5       \\
PPO clip range             & 0.2         & 0.2      & 0.2       \\
learning rate              & 0.001       & 0.0002  & 0.0002    \\
total timesteps            & 10M         & 20M      & 1M      \\
\# environments            & 16          & 8        & 1        \\
\# timesteps per rollout   & 128         & 128      & 1000       \\
\# epochs per rollout      & 4           & 4        & 4         \\
\# minibatches per rollout & 8           & 4        & 4         \\
frame stack                & 1           & 2        & 1         \\ \bottomrule
\end{tabular}
\caption{PPO training hyperparameters for three tasks.}
\label{tab:ppo}
\end{table*}

When applying our approach, we need to choose the saliency threshold to select saliency that will be penalized, weighting parameter $\alpha$ to balance selective input gradient regularization and policy distillation, learning rate and online state buffer size. Furthermore, in practice, since we use conduct input gradient regularization based on perturbation-based saliency maps collected in advance instead of producing them from the state buffer, we also need to choose the number of perturbation-based saliency maps in the offline perturbation saliency dataset. We list these hyperparameters in Table \ref{tab:digr}.

\begin{table*}[h]
\centering
\begin{tabular}{@{}cccc@{}}
\toprule
Hyperparameters                     & Red-Fetch-Green & Breakout & CARLA Driving \\ \midrule
saliency threshold                  & 0.1         & 0.1      & 0.1       \\
weighting parameter $\alpha$                & 0.01        & 0.01     & 1       \\
learning rate                       & 0.001       & 0.001    & 0.0002     \\
optimizer                       & Adam       & Adam    & RMSprop     \\
online state buffer size                   & 10K         & 10K      & 10K       \\
\# perturnation-based saliency maps & 1K          & 10K      & 2.5K      \\ \bottomrule
\end{tabular}
\caption{DIGR hyperparameters for three tasks.}
\label{tab:digr}
\end{table*}

In this work, we use multiple saliency map methods including Vanilla Gradient, Guided Backpropagation, Grad-CAM, Integrated Gradient, Smooth Gradient, Guassian-Blur Perturbation and SARFA. For Grad-CAM, we report the saliency maps extracted from the last convolutional layer. For Integrated Gradient method, we use 50 interpolation steps to calculate the saliency maps. In our experiments, Smooth Gradient saliency maps are produced by applying SmoothGrad  on Guided Backprop saliency maps. For SmoothGrad, we set the noise scale $\sigma$ as 0.15 and the number of samples as 20. One important hyperparameter in Gassuain-Blur perturbation-based method is the radius size of the perturbation. Based on the size of the state images and features, we set radius as 4, 8, 5 in Red-Fetch-Green, Breakout and CARLA Autonomous Driving. SARFA is based on Gassuain-Blur perturbation and thus share the same hyperparameters.

\section{Additional Experiment Results}
\begin{figure}
    \centering
    \includegraphics[width=0.85\textwidth]{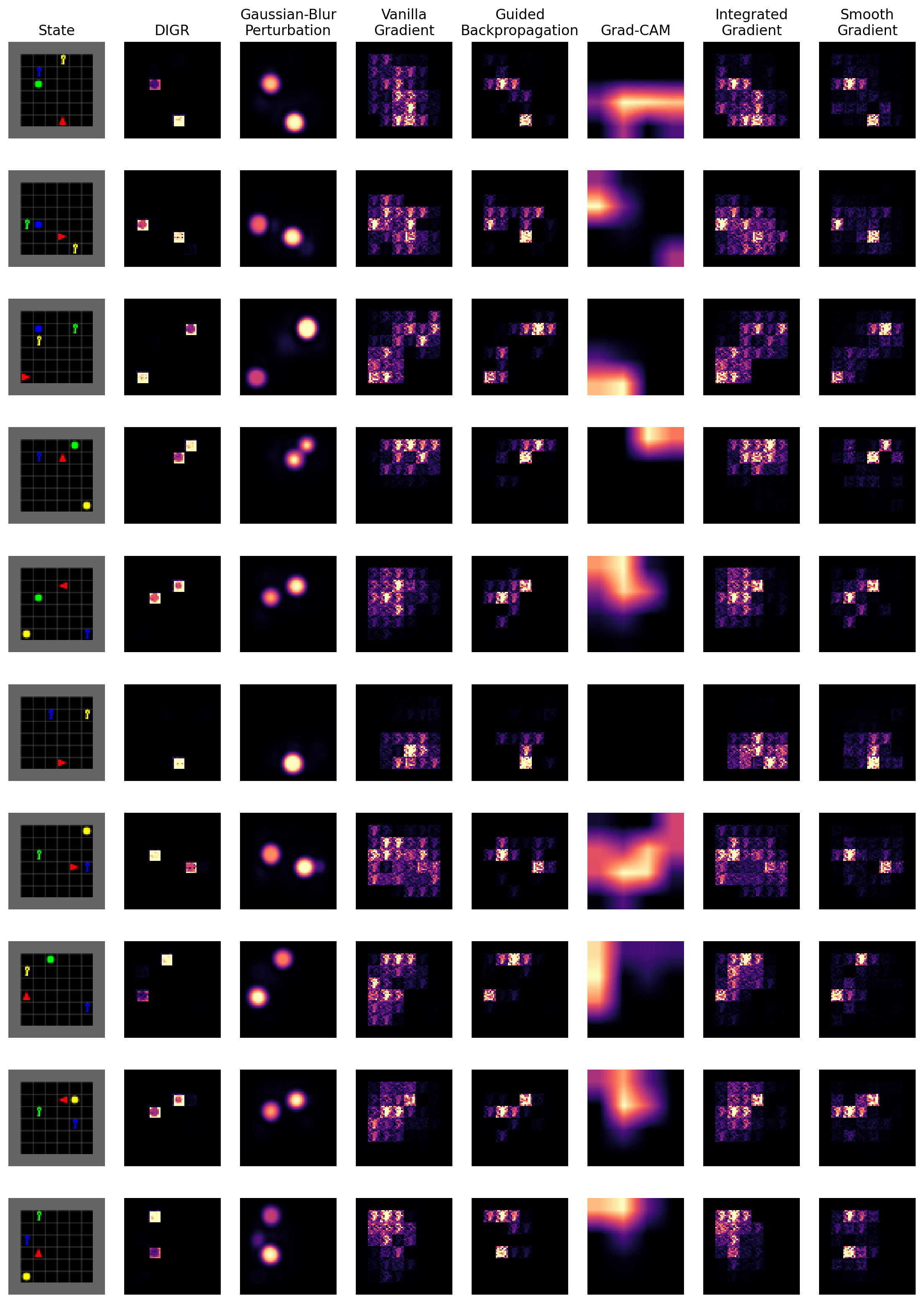}
    \caption{Supplementary saliency map examples on Red-Fetch-Green}
    \label{fig:appendix_minigrid}
\end{figure}

\begin{figure}
    \centering
    \includegraphics[width=0.85\textwidth]{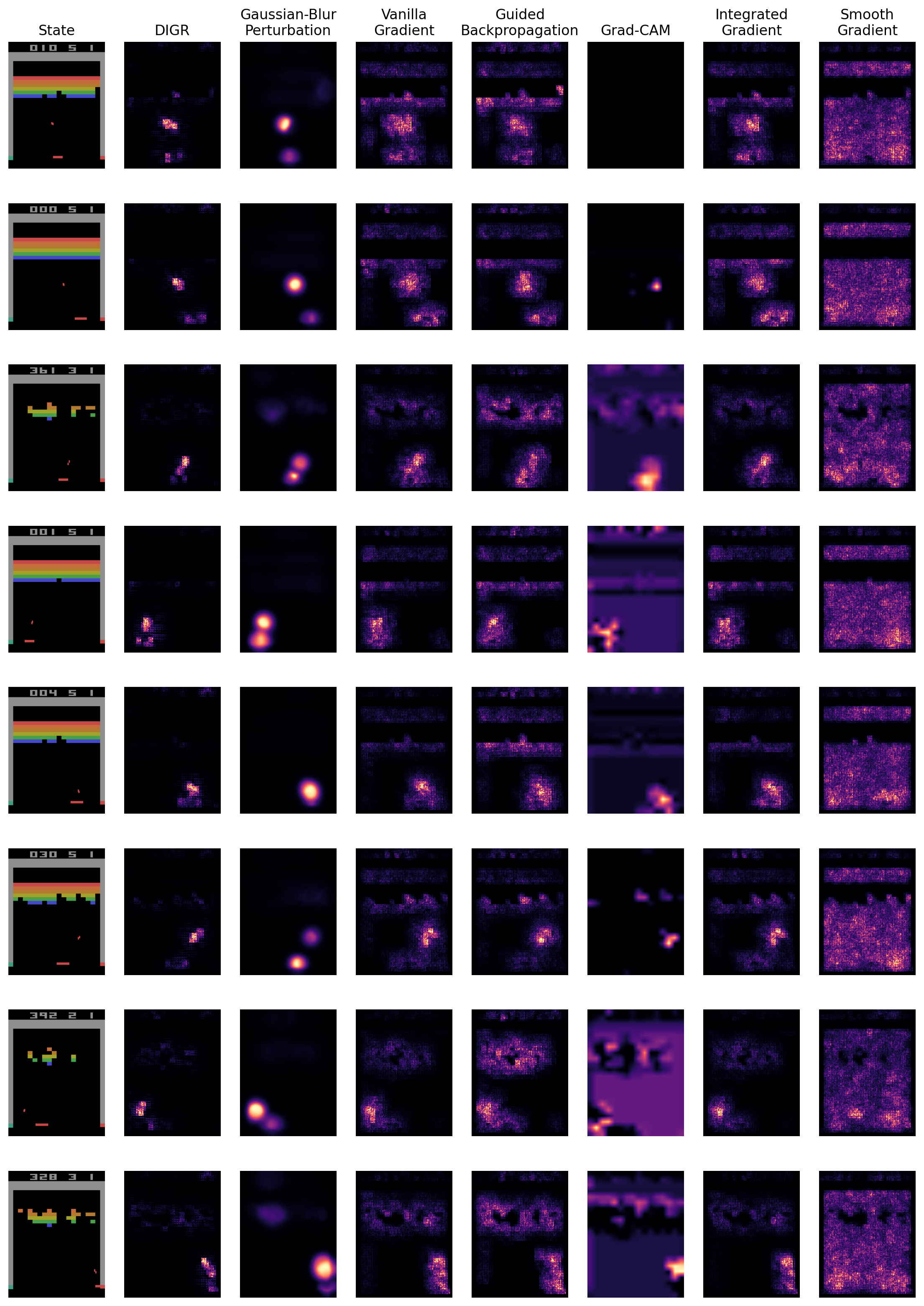}
    \caption{Supplementary saliency map examples on Breakout}
    \label{fig:appendix_breakout}
\end{figure}

\begin{figure}
    \centering
    \includegraphics[width=0.85\textwidth]{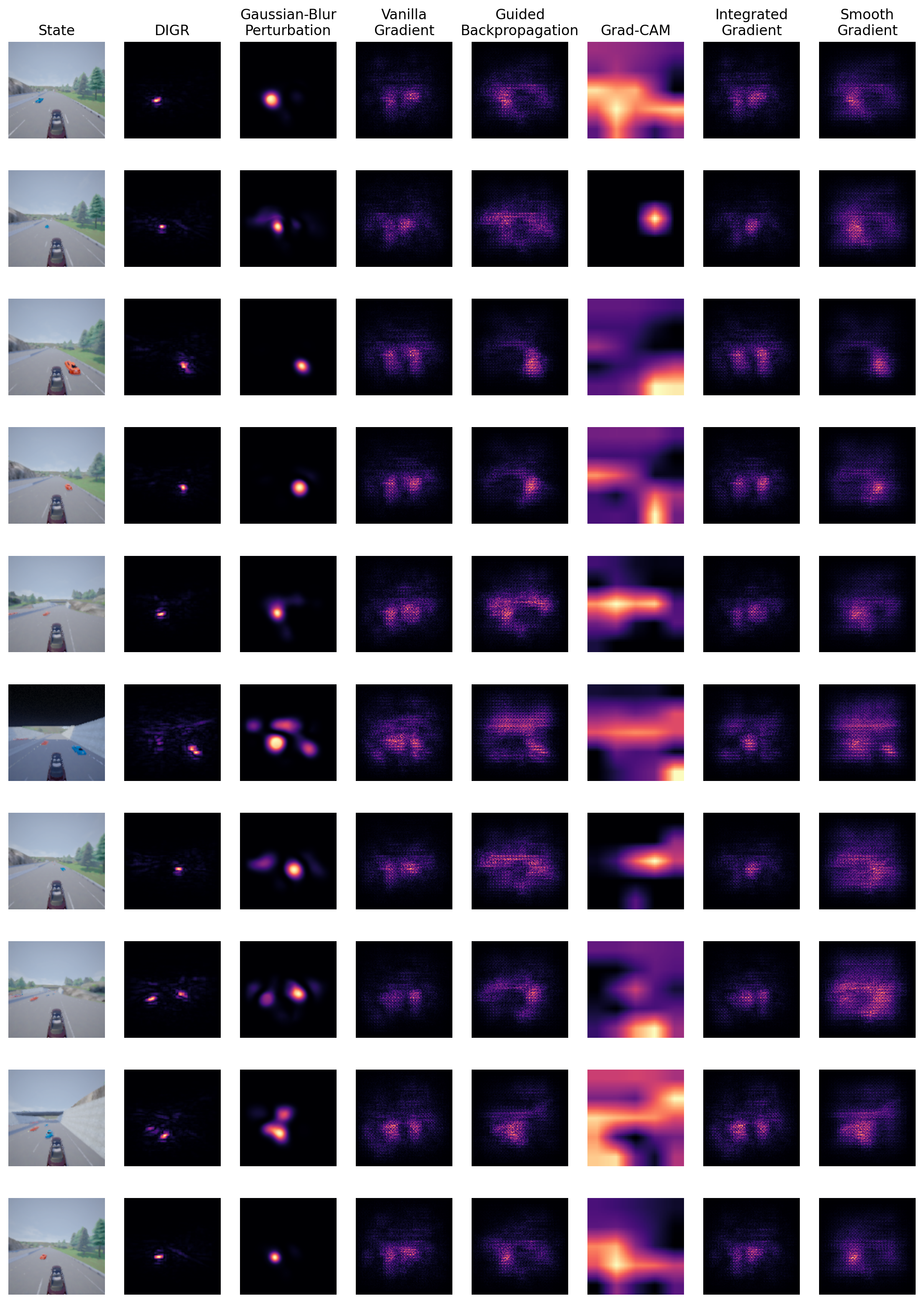}
    \caption{Supplementary saliency map examples on CARLA Autonomous Driving}
    \label{fig:appendix_carla}
\end{figure}

In this section, we provide more examples to demonstrate the effectiveness of DIGR. Besides Vanilla Gradient and Gaussian-Blur perturbation-based saliency maps, we also provide Guided Backprop, Grad-CAM, Integrated Gradient and Smooth Gradient saliency maps for comparison. All these saliency maps except DIGR are produced by the policy trained with PPO algorithm. The results on Red-Fecth-Green, Breakout and CARLA Autonomous Driving are shown in Figure \ref{fig:appendix_minigrid}, \ref{fig:appendix_breakout} and \ref{fig:appendix_carla}.

\end{document}